%% file: iscram-main.tex
\documentclass{iscram}

\usepackage[table]{xcolor}
\usepackage{graphicx}

\usepackage{float}
\usepackage{enumitem}
\usepackage{siunitx}
\usepackage{etoolbox}
\usepackage{afterpage}
\usepackage{amsmath}   
\usepackage{listings}
\usepackage[T1]{fontenc}
\usepackage{todonotes}

\usepackage{xcolor}
\usepackage[normalem]{ulem}
\usepackage{soul}
\usepackage{amsmath}

\lstdefinestyle{promptstyle}{
    basicstyle=\ttfamily\small,
    breaklines=true,
    frame=leftline,
    rulecolor=\color{gray!45},
    backgroundcolor=\color{gray!5},
    framesep=6pt,
    xleftmargin=0.35cm,
    xrightmargin=0.2cm,
}

\usepackage{xspace}
\renewcommand{\sectionautorefname}{\S\kern-0.2em}
\renewcommand{\subsectionautorefname}{\S\kern-0.2em}

\newcommand{\dataset}{\texttt{UrgencyScenarios}}

\iscramset{
    WiP Paper 2026={Social Media \& Crisis Communication, narratives, signals, and sentiments},
    title={LLM-Powered Automatic Translation and Urgency in Crisis Scenarios},
    short title={LLMs, Translation, and Urgency},
    author={
            short name={B. Ticona.},
            full name=Belu Ticona\thanks{corresponding author},
            affiliation={George Mason University\\United States\\\href{mailto:mticonao@gmu.edu}{\url{mticonao@gmu.edu}}},
          },
    author={
    full name=Antonios Anastasopoulos,
    affiliation={George Mason University\\ United States\\\href{mailto:antonis@gmu.edu}{\url{antonis@gmu.edu}}},
  },
}


\begin{document}
\maketitle
\abstract{Large language models (LLMs) are increasingly proposed for crisis preparedness and response, particularly for multilingual communication. However, their suitability for high-stakes crisis contexts remains insufficiently evaluated. This work examines the performance of state-of-the-art LLMs and machine translation systems in crisis-domain translation, with a focus on preserving urgency, a critical property for effective crisis communication and triage. Using multilingual crisis data (TICO-19, 30 languages) and a newly introduced urgency-annotated dataset of 100 scenarios translated into 29 languages, we show that dedicated translation models and LLMs exhibit substantial quality degradation, particularly for low-resource languages. Beyond translation quality, we conduct a human annotation study revealing a striking asymmetry: human assessors maintain consistent urgency judgments regardless of prompt language, while LLM-based urgency classifications vary widely across languages for identical scenarios, at times spanning the full range from \emph{Not Urgent} to \emph{Critical}. These findings highlight significant risks in deploying general-purpose language technologies for crisis triage and underscore the need for multilingual, human-centered evaluation frameworks.}

\par
\section{Introduction}

\label{sec:introduction}
Large language models (LLMs) and related automated language technologies (LTs) have attracted growing scholarly and practical interest across all stages of crisis management \parencite{lei-etal-2025-harnessing}. Beyond traditional natural language processing (NLP) tasks (such as classification, information extraction, and question answering), LLMs are increasingly capable of more complex operations, including summarization, text generation, and multistep reasoning.

These capabilities substantially broaden their potential applications in situational awareness, resource and needs coordination, risk assessment, and early warning, among others \parencite{odubolaAISocialGood2025}. Furthermore, LLMs hold promise not only for managing and processing complex, heterogeneous data streams, but also for delivering operational support to the diverse stakeholders involved across the crisis management ecosystem \parencite{xuLargeLanguageModel2025}.

However, the suitability of LLMs for multilingual crisis contexts remains largely underexamined. Most existing work draws on social media data, which is mostly available in a limited number of languages \parencite{lewis-etal-2025-tapping}. Few multilingual works that cover typologically and linguistically diverse languages focus only on specific tasks and types of crises \parencite{abdul-mageed-etal-2021-mega, imranTBCOVTwoBillion2022b}. More specifically, machine translation (MT) evaluation in the crisis domain using LLMs and translation models (e.g., \cite{lankfordLeveragingLLMsMT2024, casacuberta2024covidmt, alamerCrosslingualClassificationCrisisrelated2023}) remains scarce and often fails to cover languages from underserved communities and regions across the globe.

The rapid popularization of commercial LLMs and their availability in smaller, more affordable variants (e.g., GPT-4o mini, Gemini Flash Lite) has accelerated their widespread adoption as general-purpose systems. However, these models are predominantly instruction-tuned on English data, and their multilingual capabilities emerge incidentally from large-scale pretraining rather than from explicit multilingual design.

Furthermore, it remains unclear how well LLMs align with human judgment across dimensions such as expertise, cultural context, and domain-specific values,  a concern that is particularly critical in high-stakes domains such as crisis management, where expert knowledge is essential. In this context, human-AI collaboration emerges as a promising avenue for responsible LLM adoption, underscoring the need for robust, domain-specific, and human-centered evaluation frameworks.

In this sense, our work addresses the following research questions:

\begin{itemize}
\item[\textbf{RQ1.}] \textbf{Crosslingual Performance Variability.} How does the performance of LLMs and translation models vary on crisis-related data across linguistically diverse languages? Are there systematic trends associated with geographic region?
\item[\textbf{RQ2.}] \textbf{Human-LLM Alignment in Urgency Perception.} To what extent does LLM performance on crisis-specific tasks align with human judgment? Are LLMs consistent across languages, and how do language and cultural context influence this alignment?
\end{itemize}

To answer these questions, in the MT For Crisis Domain Section (\autoref{sec:translation}3) we evaluate LLMs and dedicated translation models using COVID-19 human-translated parallel corpora spanning 30 geographically diverse languages. We focus primarily on open-weight models, as these are more accessible and adaptable than large commercial systems, a distinction that is particularly relevant for humanitarian NGOs, local governments, and crisis responders who serve affected communities in underserved regions, often under significant resource constraints.

In the Human and LLM Perception of Urgency Section (\autoref{sec:urgency}4), we examine how LLM performance on urgency classification, a crisis-relevant task in which translation operates as an underlying process, varies across languages. We further conduct a small-scale human evaluation to assess the degree to which LLM urgency judgments align with those of human assessors in crisis contexts.

Our results show that both dedicated machine translation systems and state-of-the-art LLMs exhibit highly variable performance in the crisis domain across diverse languages. Critically, we demonstrate that translations that appear linguistically adequate can nonetheless distort the conveyed level of urgency, and that urgency judgments produced by LLMs vary substantially depending on the language of the prompt and input content.

\begin{figure}[b!]
\centering
\includegraphics[trim={44.5cm 55cm 51.5cm 1cm}, clip, width=1\textwidth]{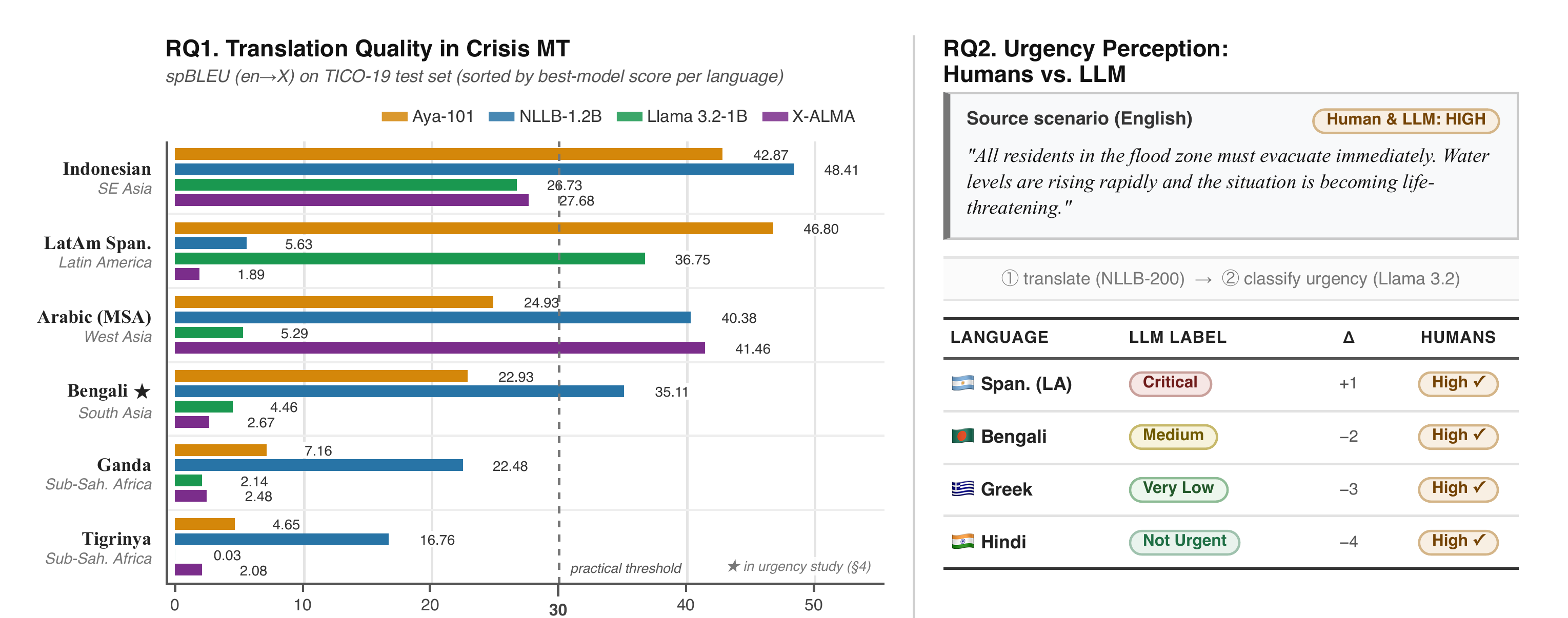}
\caption{\textbf{Left (RQ1):} spBLEU scores ($en{\rightarrow}X$) on TICO-19 crisis data across four model families. All models collapse for Sub-Saharan African languages (Ganda, Tigrinya); Latin Am.\ Spanish exposes extreme inter-model inconsistency (Aya-101: 46.80 vs.\ NLLB-1.2B: 5.63). The dashed line marks the practical threshold (${\geq}30$). \textbf{Right (RQ2):} Urgency classification of a \dataset{} scenario translated into three languages via NLLB-200 and classified by Llama~3.2 across 30 languages. Human annotators agree across all languages (High); LLM labels span four distinct categories for the same source sentence, ranging from \textit{Not Urgent} to \textit{Critical}.}
\label{fig:overview}
\end{figure}

These findings raise serious concerns about the consistency and reliability of LLMs for crisis-relevant tasks, not only in multilingual settings, but more broadly as systems whose alignment with human judgment remains poorly understood. This is especially consequential for crisis practitioners and humanitarian actors seeking to deploy such systems, even within human-AI collaboration frameworks, to serve affected populations globally.

Our work underscore the risks of adopting general-purpose LLMs in high-stakes crisis environments without rigorous, domain-specific evaluation. We therefore advocate for multilingual, human-centered evaluation of LLMs on crisis tasks, and for deployment practices grounded in the operational realities of emergency response. Figure~\ref{fig:overview} provides a joint preview of our main findings: translation quality collapses for low-resource languages (RQ1), and LLM urgency labels are highly unstable across languages for the same source scenario while human assessments remain consistent (RQ2).

\section{Related Work: Language Technologies for Crisis Response}
\label{sec:related}

\paragraph{Automatic Translation for Crises}
MT for crisis has progressed from early rule-based approaches, such as the Apertium initiative for Kurdish languages \parencite{globalTranslatorsBordersDevelops2016, forcadaApertiumFreeOpen2016}, to the rapid-response statistical MT frameworks established during the 2010 Haiti earthquake \parencite{lewisHaitianCreoleHow2010}. This led to large-scale, cross-institutional programs like DARPA LORELEI, which advanced neural MT specifically for health-related crisis communication in languages such as Arabic and Swahili \parencite{strasselLORELEILanguagePacks2016, traceyCorpusBuildingLow2019}.

The COVID-19 crisis motivated numerous parallel data collection initiatives across multiple languages \parencite{casacuberta2024covidmt, anastasopoulosTICO19TranslationInitiative2020}, producing some of the few crisis-domain parallel corpora available to date. However, fine-tuning efforts built on this data have largely focused on a small number of European languages \parencite{wayRapidDevelopmentCompetitive2020, roussisBuildingEndtoEndNeural, lankfordMachineTranslationCovid2024}.

More recent studies have explored the potential of LLMs for crisis MT in low-resource languages, comparing vanilla and fine-tuned versions of commercial LLMs with MT models specifically developed for low-resource settings, such as NLLB \parencite{teamNoLanguageLeft2022c}. For instance, \citeauthor{lankfordLeveragingLLMsMT2024} show that fine-tuned NLLB on in-domain data outperforms custom versions of GPT-4 for Marathi and Irish.

Concurrent with our work, \textcite{merx-etal-2025-openwho} introduce OpenWHO, a parallel corpus drawn from the World Health Organization's e-learning platform, and evaluate a set of MT and LLM models including NLLB-54B, Gemini, DeepSeek-v3, and Gemma~3. While this represents a meaningful contribution to health-domain MT, the low-resource subset covers 9 languages with approximately 200 parallel sentences each, and the evaluation focuses on large commercial models.   

Our work is complementary in this regard, offering a broader multilingual scope across the 30 typologically diverse languages evaluated from TICO-19, with a focus on open-weight models more accessible to the humanitarian organizations and low-resource institutions most likely to serve crisis-affected communities. Similarly, \textcite{bansalMTReadyNext2026} also study MT on the same dataset, focusing on commercial models from the main MT and LLM providers, such as Google, Microsoft, Open AI. 

Collectively, these efforts highlight meaningful progress in crisis MT, yet parallel corpora for other crisis types remain absent from the literature, and coverage of typologically diverse and low-resource languages is still limited. 




\paragraph{Urgency Assessment}
Among the applications of LLMs in crisis settings, automatically assessing the urgency of incoming information has emerged as a key task for disaster situation assessment and response prioritization \parencite{lei-etal-2025-harnessing}. Studies have shown that LLM-based urgency assessment can match untrained emergency health personnel but still fall short of professionally trained doctors, such as emergency department staff \parencite{masanneckTriagePerformanceLarge2024}.

Moreover, \citeauthor{leePerformanceChatGPTGemini2025} evaluated commercial LLMs in real-life clinical conversations, showing that they can accurately classify the urgency of emergency department patients. However, these evaluations are conducted solely in English, lacking a multilingual assessment. \citeauthor{khullarScriptGapEvaluating2025} tackle this gap by evaluating LLMs (such as GPT-4-o, Claude 4.5, and Qwen3) in 6 South Asian languages, studying the effect of orthographic variations on real-world data in the health domain. Their results show a degradation in performance for Romanized messages, revealing a blind spot in LLM-based urgency classification systems. Our work expands the scope of prior work by broadening the range of languages covered and including open-weight LLMs. 

\section{MT for Crisis Domain: The COVID-19 Case Study}
\label{sec:translation}

In this section, we evaluate the performance of language models in MT in languages linguistically and geographically diverse from different regions of the world. Our goal is to evaluate these models specifically in the \textit{crisis domain}, since their performance generally depends on the amount and type of data in which they are trained on.

We understand by \textit{crisis domain} all tasks that involve the process and management of data related to crises, covering the wide range of possible definitions depending on the theoretical frameworks used\footnote{In crisis literature, \textit{crisis} can be defined from an event-based \parencite{lerbinger2012crisis}, attribution-based \parencite{coombs2010handbook}, and temporal-dynamics \parencite{paraskevas2013mitroff} perspective, among others approaches.}.  

In this work, we focus our study in public health crisis, using COVID-19 data as a representative case. The scale and globally implications of this event motivated the creation of parallel corpora in multiligual settings, which makes it more suitable for our evaluation. Furthermore, this is one of the only crises type for which parallel corpora is available across a wide range of languages globally.

With regard to the models, our evaluation primarily considers open-weight models, as their accessibility and adaptability make them more suitable for the humanitarian NGOs, local governments, and crisis responders most likely to serve affected communities in underserved regions. Beyond adaptability through post-training techniques, open-weight models can be deployed locally, a critical property in disaster zones where internet connectivity is unreliable, and offer more predictable cost structures compatible with humanitarian budget cycles. They are also auditable, allowing organizations to inspect and certify model behavior before deployment, a requirement increasingly demanded by institutional donors and accountability frameworks.\footnote{Additional advantages over commercial systems include: reproducibility (fixed weights yield stable outputs, unlike silently updated APIs); resilience to model discontinuation; and data sovereignty, as routing sensitive crisis data through third-party APIs may conflict with data protection regulations or community trust norms. We acknowledge, however, that open-weight models also carry limitations: for example, they require technical personnel for deployment and adaptation, and are less immediately accessible than off-the-shelf solutions (see Limitations (\autoref{ssec:limitation-rq1})).} Accordingly, our evaluation covers languages less-represented in the Internet, focusing on Africa and Asia.


\subsection{Methodology}
\label{subsec:mtexp}

\paragraph{Data}
We use the TICO-19 dataset in our evaluations, a publicly available crisis-domain parallel corpus covering linguistically and geographically diverse languages from Africa and Asia mainly, as well as few languages from Latin America and Europe \parencite{anastasopoulosTICO19TranslationInitiative2020}. This dataset encompasses a range of sources, including Wikipedia articles, PubMed publications, news reports, and NGO communications, human-translated from English into 39 low- and medium-resource languages. For our experiments, we use the test set, which contains approximately 2,000 sentences per language, comprising a total of 70,000 sentences.

\paragraph{Models}
We evaluate a set of open-weight models spanning two complementary paradigms: dedicated multilingual MT systems and general-purpose LLMs. This selection reflects both the current landscape of low-resource MT and the growing role of LLMs in translation tasks for specialized domains such as crisis response. The official language support of each model varies substantially, thus to facilitate the analysis we provide the list the TICO-19 evaluation languages alongside their official support status per model, which contextualizes the performance differences discussed in the analysis (see Appendix~\autoref{app:lang-support}). 

We include \texttt{NLLB-200} \parencite{teamNoLanguageLeft2022b} as a state-of-the-art multilingual MT system built explicitly for low-resource languages; \texttt{X-ALMA} \parencite{xuXALMAPlugPlay2025} as an LLM-based MT model with a modular multilingual architecture; \texttt{Aya-101}  \parencite{ustun-etal-2024-aya} as a multilingual LLM trained with broad language diversity as a core design objective; and \texttt{Llama}~3.2 \parencite{grattafioriLlama3Herd2024} as a widely adopted general-purpose LLM that, despite limited official multilingual support, serves as a common backbone for domain- and language-specific fine-tuning.

The evaluations were conducted in June 2025 for \texttt{NLLB-200} and \texttt{Aya-101}, and October 2025 for \texttt{X-ALMA} and \texttt{Llama}~3.2. We report the exact versions used for each model below.

\texttt{NLLB-200} is a suite of neural MT models covering more than 200 languages, developed by Meta as part of the \textit{No Language Left Behind} initiative, with an explicit focus on lesser-resourced languages underserved by existing MT systems. As a dedicated multilingual MT system, it constitutes a natural reference point for low-resource translation evaluation, and is widely used as a backbone for fine-tuned MT systems targeting specific low-resource language families, such as indigenous languages of the Americas \parencite{de-gibert-etal-2025-findings} and languages of India \parencite{pakray-etal-2025-findings}. We use the open-weight variant released through the Open Neural Machine Translation project.\footnote{\url{https://github.com/gordicaleksa/Open-NLLB}}

\texttt{Aya-101} is a multilingual generative model developed by Cohere Labs, supporting 101 languages of which approximately 50\% are considered low-resourced \parencite{ustun-etal-2024-aya}. Unlike most multilingual instruction-following models, \texttt{Aya-101} was explicitly designed to counteract English-centricity in training data, with only 21.5\% of its instruction data in English. We include it as a representative multilingual LLM with broad cross-lingual coverage and a strong orientation toward underrepresented languages. We use the official open-weight model available on Hugging Face.\footnote{\url{https://huggingface.co/CohereLabs/aya-101}}

\texttt{Llama}~3.2 is a family of multilingual generative models, available in 1B and 3B parameter sizes, released by Meta in both pretrained and instruction-tuned variants \parencite{grattafioriLlama3Herd2024}. Although the model officially supports only eight high-resource languages, it is widely adopted across applied NLP tasks due to its open-weight availability and strong general-purpose performance, making it a common backbone for language- and domain-specific fine-tuning. We include Llama~3.2 to assess the zero-shot translation capabilities of a general-purpose LLM not explicitly optimised for low-resource or multilingual scenarios, providing a contrasting reference point to the other models in our evaluation. We used \texttt{Llama-3.2-1B-Instruct} through vLLM \parencite{kwon2023efficient} for our experiments.

\texttt{X-ALMA} is an open-weight LLM-based MT model with a modular architecture that groups linguistically similar languages into dedicated modules, officially supporting 50 languages \parencite{xuXALMAPlugPlay2025}. Its plug-and-play design enables language-specific components to be integrated or extended without retraining the full model, making it well-suited for adaptation to new languages within an existing linguistic family. For TICO-19 languages outside the 50 officially supported, we assign the closest available language group in the training set based on linguistic proximity (see Table~\ref{tab:lang-support}). We include \texttt{X-ALMA} as a representative of the emerging class of LLM-based MT systems that combine the generative capabilities of large language models with translation-specific training objectives. We use the official release available on GitHub.\footnote{\url{https://github.com/fe1ixxu/ALMA}}

\paragraph{Metrics}
We evaluate translation performance using SacreBleu \parencite{post-2018-call}, a well-known MT evaluation framework that computes automatic metrics based on the lexical distance between the hypothesis sentence and its reference translation. We report two complementary metrics: 1) spBLEU, a variation of the BLEU score that operates on sub-words units to better handle diverse scripts \parencite{goyal-etal-2022-flores}, and 2) chrF++ \parencite{popovic-2017-chrf}, which improves on upon standard-level evaluation by incorporating word n-grams to increase robustness for morphological variations. We limit our evaluation to string metrics to assess all languages equally; however, language-specific metrics that have been trained on data for the languages evaluated could be considered (see Limitation \autoref{ssec:limitation-rq1}3).

\subsection*{Analysis of Crosslingual Performance}
\label{subsec:mtresults}

\begin{description}
    \item[RQ1. Crosslingual performance variability.] How does the performance of LLMs and translation models vary on crisis-related data across linguistically diverse languages, and what systematic trends emerge regarding directionality and geographic regions?
\end{description}

\begin{figure*}[!b]
\centering
\includegraphics[width=.9\textwidth]{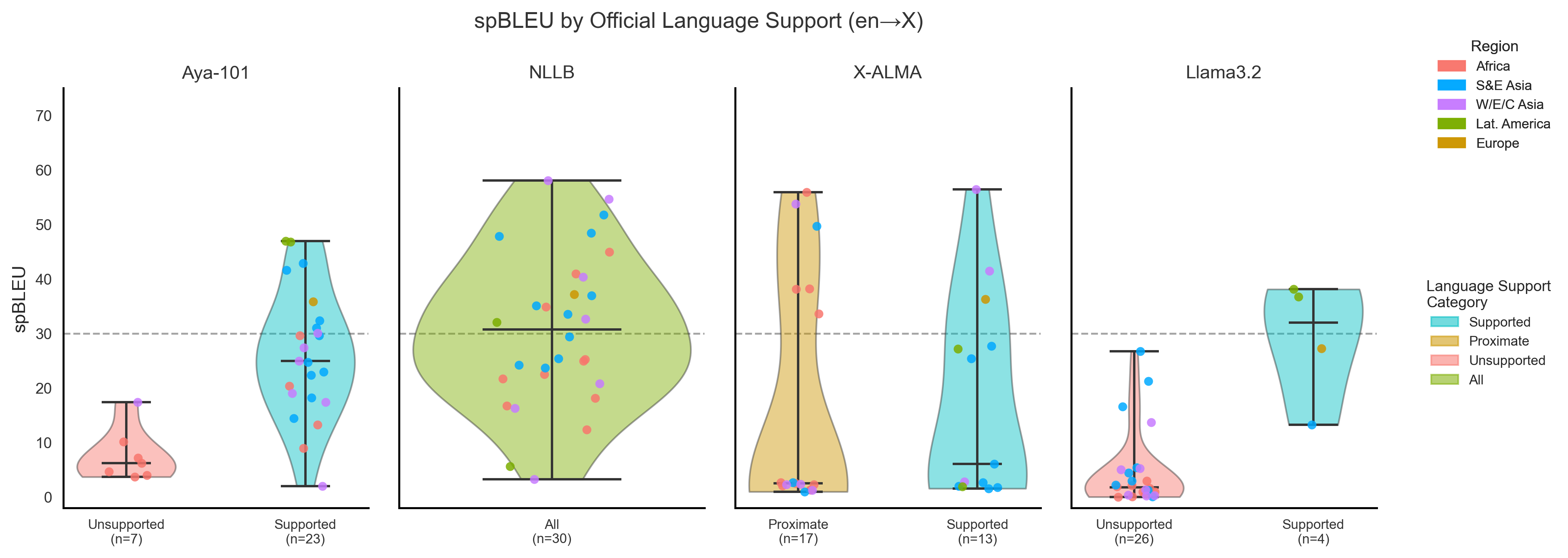}
\caption{
spBLEU performance by official language support category for each MT system in the $en \rightarrow X$ direction. Each point represents one language and is colored by geographic region. The dashed line at spBLEU~=~30 marks the approximate threshold for preserving usable semantic content. Supported languages generally achieve substantially higher performance, while unsupported or proximate languages frequently fall below this threshold. Compared to other models, \texttt{NLLB-200} exhibits more consistent performance across regions and fewer catastrophic failures.
}
\label{fig:support_violin}
\end{figure*}

In this section, we analyze machine translation (MT) performance across 30 crisis-relevant languages spanning Africa, Asia, Latin America, and Europe. Our results reveal that translation quality is highly uneven across languages and strongly associated with the degree of official language support provided by each model.

Across models, we observe a systematic \textbf{directionality asymmetry}: translation quality from English into other languages ($en \rightarrow X$) is consistently lower than translation into English ($X \rightarrow en$). This trend is particularly concerning in humanitarian settings because it suggests that current MT systems are substantially more reliable for extracting information from affected populations than for disseminating critical instructions back to those communities.

To facilitate interpretability, we apply a practical heuristic for spBLEU scores: values below $20$ indicate substantial information loss; values between $20$--$29$ convey partial meaning but contain major semantic or grammatical degradation; and values above $35$ generally correspond to understandable translations. For operational crisis settings, we consider scores above $40$ more appropriate for situational awareness and scores above $50$ suitable for broader communication tasks.

Figure~\ref{fig:support_violin} shows that official language support is strongly associated with translation quality. For \texttt{Aya-101} and \texttt{Llama}~3.2, supported languages cluster substantially above unsupported ones, while unsupported languages frequently remain below the spBLEU~=~30 interpretability threshold. This pattern is especially pronounced for \texttt{Llama}~3.2, whose performance is concentrated almost entirely on the four officially supported languages present in our benchmark (French, Brazilian Portuguese, Hindi, and Latin American Spanish). Outside these languages, performance collapses toward near-zero translation quality.

\texttt{X-ALMA} exhibits particularly high volatility. While the model achieves competitive performance for some supported or proximate languages, its distribution is strongly bimodal: several languages achieve high spBLEU scores while many others cluster near complete failure. This instability suggests that aggregate multilingual performance can obscure substantial disparities between linguistic communities.

In contrast, \texttt{NLLB-200} demonstrates the most geographically stable behavior. Although it does not always achieve the highest score for every language, it maintains comparatively consistent performance across regions and avoids the catastrophic degradation observed in the other systems. The narrower and more continuous distribution visible in Figure~\ref{fig:support_violin} suggests that broad language coverage may be more operationally reliable than strong performance concentrated on a limited subset of languages.

We further observe strong regional disparities. Languages from Africa and West/Central Asia are disproportionately represented among the lowest-performing cases across models. For example, several systems obtain spBLEU scores below $15$ for Ganda, Dari, Tigrinya, and Sorani Kurdish in the $en \rightarrow X$ direction. In many of these cases, outputs contain severe semantic distortion or near-total information loss.

These findings suggest that average multilingual benchmark performance can obscure substantial regional inequities. A model that appears effective for one linguistic community may fail catastrophically for another, despite both being represented within the same ``multilingual'' system. Consequently, deployment decisions in humanitarian contexts should not rely solely on aggregate multilingual claims, but instead require preliminary evaluation on the specific linguistic varieties used by affected populations.

Most concerningly, for several languages spoken in conflict-prone or disaster-relevant regions, all evaluated models exhibit substantial degradation. For example, the highest observed spBLEU for Tigrinya is only $16.76$, while Sorani Kurdish reaches at most $20.82$ in the evaluated configurations. Under our interpretability criteria, these scores indicate that current MT systems remain unreliable for safety-critical communication in these languages. In such settings, human translation and local linguistic expertise remain essential for safe crisis response.
\subsection{Limitations}
\label{ssec:limitation-rq1}
Our evaluation is limited in the number of languages and models covered, which is mainly due to the limited existing human-translated parallel corpora in multiple languages to conduct a comparative analysis. Our work focuses only on vanilla versions\footnote{Models without adaptation to specific-domain data (e.g. without fine-tuning)} of translation models commonly used in academic spaces as based models for domain and language/region specific versions (e.g. \textcite{sanchezLargeLanguageModels2025}). Further improvement could be achieved if they were fine-tuned in crisis health data.

Moreover, our evaluation only covers two of the most traditional automatic string metrics, leaving others for further studies. Future work could focus on language-specific evaluations, covering specific metrics according to the language and translation direction. For instance, \textcite{li-etal-2025-ssa} introduce SSA-COMET which outperforms other semantic metrics thanks to metric training on large-scale human annotated MT evaluations in African languages. Furthermore, other work analyze limitations of semantic metrics when used in unseen language in training  \parencite{zebaze-etal-2025-context, zouhar-etal-2024-pitfalls}.

Other limitation of our work is that we only explore a zero-shot prompt technique and models are known for their high variability to prompting techniques. A future improvement could explore multiple executions and example selection, which have especially shown better results in the high-to-low resource translation direction \parencite{zebaze-etal-2025-context}.

%



\subsection{Discussion and Future Work}

In summary, while models specialized in language diversity and translation show promise for local situational monitoring, our results confirm that they remain unreliable for crisis communication dissemination. Deploying off the shelf models in highly specialized domain scenarios requires dedicated assessment on the local languages of the deployment context, as performance inconsistencies can lead to the spread of life threatening misinformation, the deterioration of institutional trust, or the exclusion of linguistic communities from essential aid.

These findings align with broader trends in low resource machine translation, where fine tuned systems remain the strongest baseline across diverse regions, including Indigenous languages of the Americas \parencite{de-gibert-etal-2025-findings}, low resource Indic languages \parencite{pakray-etal-2025-findings}, and African languages, where even commercial LLMs fall short of fine tuned baselines \parencite{ojo2025afrobenchgoodlargelanguage}.

In the crisis domain, however, such fine tuning remains out of reach for most languages, since parallel crisis corpora are scarce and the infrastructure needed to build and evaluate domain adapted systems does not exist for many of the languages most exposed to humanitarian emergencies. Our evaluation of vanilla models therefore reflects a realistic picture of what is currently deployable in crisis affected regions using open-weight models.

Future work could extend this evaluation to include commercial and smaller instruction tuned models \parencite{ojo2025afrobenchgoodlargelanguage}, as well as more suitable metrics depending on the language. For instance, while some languages could be better assessed using semantic metrics due to the availability of MT human evaluation data, others could be better studied using aggregate approaches based on lexical metrics as used in the present work \parencite{cavalinSentencelevelAggregationLexical2025}.

\input{tico-19-mt-table}

More broadly, most models are trained on general-domain data that does not reflect the linguistic and situational characteristics of crisis scenarios, and even state-of-the-art systems struggle with robustness to non-standard input and domain-specific terminology \parencite{kocmi-etal-2025-findings}. In crisis settings, subtle changes in wording could alter risk perception, delay response, or erode trust, making it essential to go beyond translation accuracy and assess the faithful preservation of communicative intent.

This points to the need for human-centric evaluation frameworks that examine how practitioners actually rely on language technology to perform their tasks \parencite{carpuat-etal-2025-interdisciplinary}, a point we reinforce in the following section, where we study LLMs in urgency classification when translation is an underlying task.

\section{Human and LLM Perceptions of Urgency}
\label{sec:urgency}

In this section, we investigate how translation quality influences crisis communication when LLMs are integrated into crisis-related applications. While \textcite{lei-etal-2025-harnessing} analyze LLM utility across diverse tasks, including sentiment analysis, vulnerability detection, and situational assessment, as well as generative tasks like the automatic creation of evacuation plans, our work focuses on classification as a primary approach.

Classification represents a lower-complexity task that allows for a direct comparison between human and LLM distribution patterns, avoiding the high degree of nuance and discrepancy often found in generative outputs.

Specifically, we center our analysis on the dimension of \textit{urgency}. We explore the similarities and differences in how humans and LLMs perceive urgency when exposed to content conveying varying levels of gravity.

Although this study is limited to a single dimension, this methodology is extensible to other dimensions such as severity, factuality, or sentiment, among others. Through this approach, we aim to determine how translation quality impacts the conveyance of meaning for critical \textit{proxies} of the semantic dimensions essential to effective crisis translation.

\subsection{Methodology}

There exist limited human-translated parallel corpora to conduct fair evaluation in crisis domain, as discussed in the previous section. In this sense, our approach explores the use of LLMs to create synthetic data that span a wide level of urgency.

We expect real-world data to reflect even more complex nuances on urgency assessment, and therefore, being a more complicated task that the one addressed in this work. In other words, if LLMs' performance on urgency assessment using synthetic data shows poor alignment with humans, the use of real-word crisis data would perish even more this alignment. 

\subsubsection{The \dataset{} Dataset}
\label{subsec:dataset}
Due to the limited availability of human-translated multilingual data, we create our own dataset in English, automatically translating it into additional 29 languages. We use ChatGPT-4o-mini (OpenAI, accessed October 2025\footnote{\url{https://chatgpt.com}}) to iteratively generate scenarios conveying 6 different levels of urgency.

We prompt to cover scenarios related to natural and climate disasters, manually selecting those that described new scenarios than the ones already generated in previous iterations. This observation aligns with our assumptions that LLMs tend to generate similar content, conveying less variety of urgency (see discussion in Limitations).

In total, we manually selected and curated 100 sentences, each of them describing a scenario with a level of urgency in the context of natural disaster. We translated this set into 29 medium and low-resource languages using a small version of \texttt{NLLB-200} \parencite{teamNoLanguageLeft2022c}, creating a multilingual set of 3,000 scenarios; we refer to this dataset as \dataset.


\label{subsec:mturgency}

\paragraph{Annotation Setup}
First, we conduct a small-scale human study to understand how urgency perception fluctuates across languages in assessing crisis scenarios. We focus on graduate students whose first language is the language of study and who professionally work in the United States in English-speaking environments, demonstrating a proficient level of English. We recruit two native speakers for each of the following four languages: Spanish, Greek, Bengali, and Hindi. Note that Greek is not part of the TICO-19 evaluation set used in the MT For Crisis Domain Section (\autoref{sec:translation}). 


We split \dataset{} into two, providing half set in English and the other half in the annotator's native language. In order to have the dataset fully annotated in all languages, we provide separate halves to each of the speakers of the same language, producing a single annotation per language per sentence.

We ask the annotators to classify each scenario choosing among six urgency categories: Non-critical, Very-Low, Low, Medium, High, Critical; along with an example for each one in a few-shot configuration (see prompt in the Appendix (\autoref{app:promptfull})). In this way, we obtained a set of annotations per language, and four sets in English.

Then, we conduct LLM annotations using \texttt{Llama}~3.2 as a representative model. We provide the whole translated prompts, as was done with the human annotations. Due to the generative nature of the model's response, we capture model's urgency annotation for each scenario by analyzing the inclusion of the label provided in the prompt.\footnote{A further approach could semantically analyze the LLMs annotation, instead of simply pattern matching with the given label.}

\subsection*{Results and Analysis: Human Alignment and Urgency Case Study}
\label{subsec:llmurgency}

\begin{description}
    \item[RQ2. Human Alignment: Urgency Case Study] To what extent does LLM performance on crisis-specific tasks align with human judgments? Are LLMs consistent across languages? How do language and cultural context influence this alignment?
\end{description}

\begin{figure}[!ht]
\centering

\includegraphics[trim={-0.15cm 0cm 0 -0.2cm}, clip, width=0.33\textwidth]{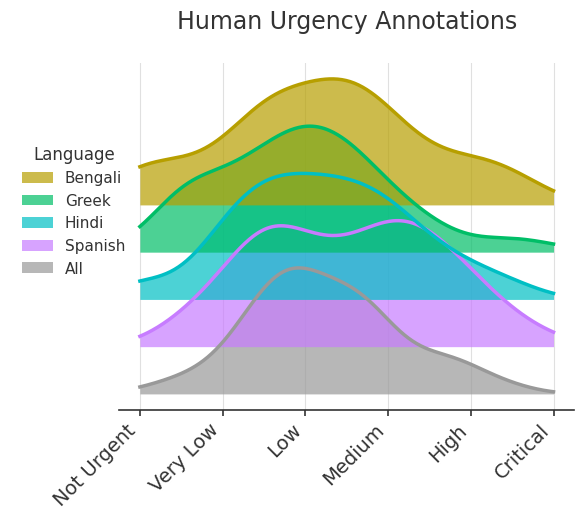}
\includegraphics[trim={0.1cm 0 0 0}, clip, width=0.6\textwidth]{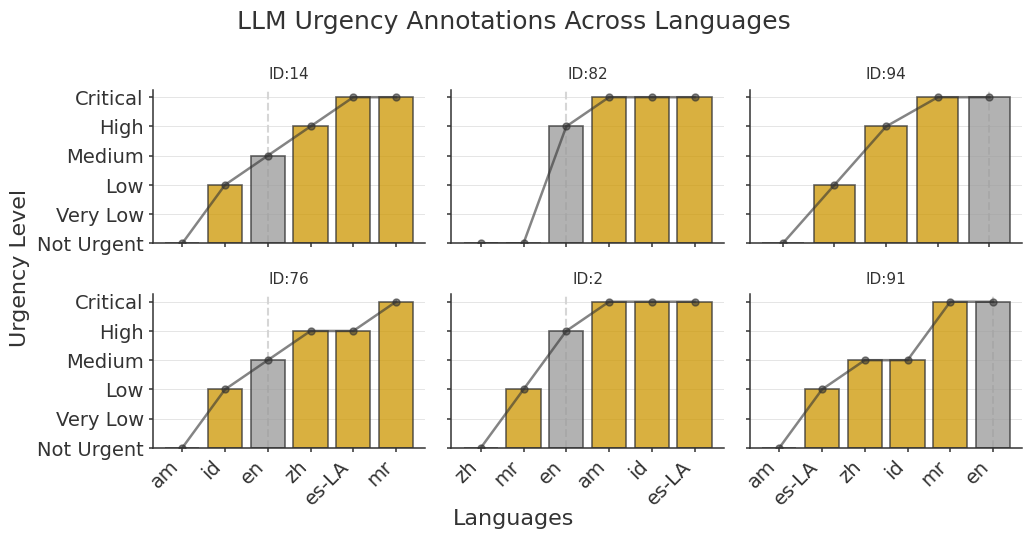}
\caption{Distribution of urgency scores per language group across all annotated sentences, whether in English or the annotator's native language (left). Cross-lingual LLM urgency assessment for the same scenario: the model assigns different labels depending on the prompt language (right). While human annotators are overall consistent, the LLM shows substantial cross-lingual instability.}
\label{fig:humanstudy}
\end{figure}

\paragraph{Human annotation consistency.}
In response to \textbf{RQ2}, our results indicate that human participants, regardless of language or cultural background, maintain a high level of consensus on urgency assessments. In contrast, the LLM utilized as an annotator demonstrates significant inconsistency when classifying the same scenario across different languages (see Figure~\ref{fig:humanstudy}).

The human annotation distributions for the evaluated languages overlap substantially, with all groups concentrating on similar urgency levels (see Figure~\ref{fig:humanstudy}, left).

\begin{figure}[!ht]
\centering
\includegraphics[trim={1.8cm 0.81cm 1.5cm 0.31cm}, clip, width=1\textwidth]{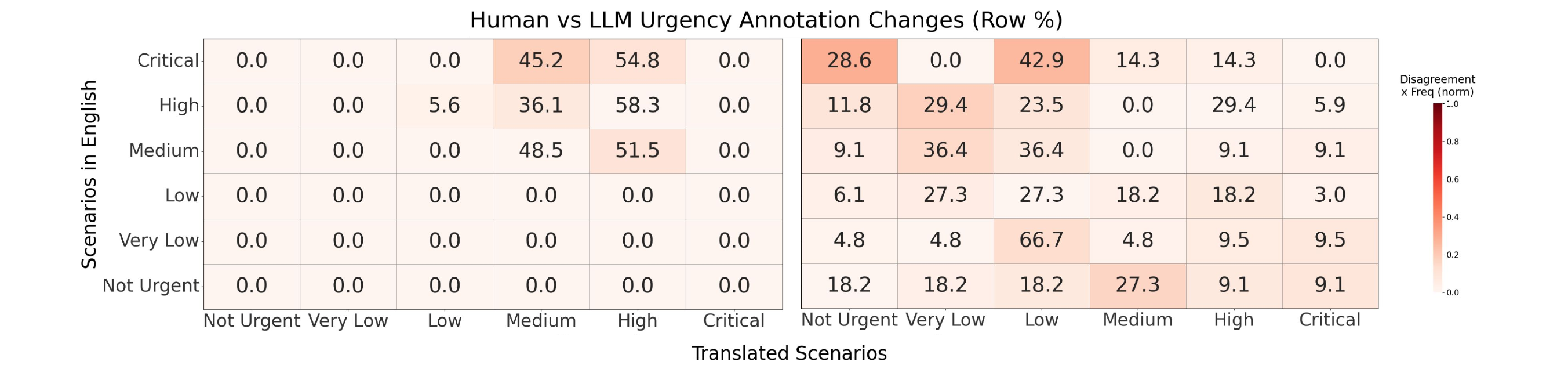}
\caption{LLMs change their urgency assessment across our scenarios, largely due to translation quality.}
\label{fig:combinedChangesHALLM}
\end{figure}

Figure~\ref{fig:combinedChangesHALLM} (left) further examines how urgency labels shift between English and translated versions across the four annotated languages (Spanish, Greek, Bengali, and Hindi). Rows correspond to labels assigned in English, while columns correspond to labels assigned in translation. Cell values indicate the percentage of label-shifts (from label assigned in English to label assigned in a non-English language), while darker colors reflect larger and more severe label shifts (see Appendix (\autoref{app:transition-matrices}) for details).

Overall, the human matrix remains strongly concentrated along the diagonal, indicating that disagreements are relatively rare and typically involve adjacent categories. For example, $66.7\%$ of scenarios labelled ``Very Low'' in English are annotated as ``Low'' in translation. Although we observe isolated severe transitions, such as ``Critical'' shifting to ``Not Urgent'', these cases are uncommon. We hypothesize that translation quality of urgency-bearing terms, such as ``immediately,'' may contribute to these outliers.

Taken together, the harmonized distributions and strong diagonal agreement suggest that humans share a broadly consistent understanding of urgency across languages, despite minor variation.

\begin{table}[!ht]
  \centering
  \caption{Cross-lingual urgency labels assigned by \texttt{Llama}~3.2 to the same gas leak scenario across six languages. Despite being \emph{Critical} in English and Indonesian, the same text is classified as \emph{Low} in Marathi and \emph{Not Urgent} in Amharic. See Appendix~\ref{app:divergence} for further examples.}
  \includegraphics[trim={35cm 48cm 35cm 1cm}, clip, width=\textwidth]{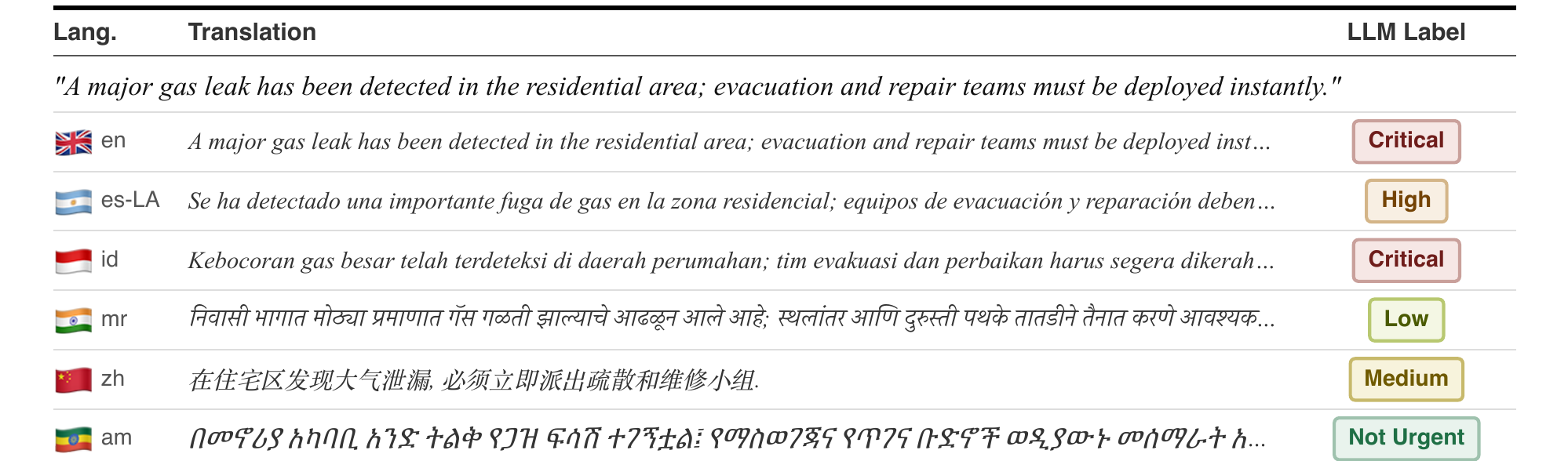}

  \label{tab:llm-divergence-short}
\end{table}


\paragraph{LLM cross-lingual instability.}
Conversely, our findings reveal substantial cross-lingual instability in LLM-based annotations. Given an identical scenario, the LLM assigns disparate urgency levels based solely on the prompt language (see Figure~\ref{fig:humanstudy}, right). The examples in Figure~\ref{fig:overview} and Table~\ref{tab:llm-divergence-short} illustrate this divergence: a gas leak scenario classified as \emph{Critical} in English is downgraded to \emph{Low} in Marathi and \emph{Not Urgent} in Amharic by the same model. Additional examples are provided in Appendix~\autoref{app:divergence}.

Furthermore, the LLM occasionally refuses to classify the input, hallucinates, or produces responses outside the provided categories. Out of the 30 languages evaluated, only 14 achieved a classification rate of at least $80\%$ across all scenarios.

Figure~\ref{fig:combinedChangesHALLM} (right) shows the corresponding transition matrix for the LLM predictions across the qualifying languages. Compared to the human matrix, the substantially darker off-diagonal regions indicate both more frequent and more severe label shifts. In particular, scenarios labelled ``Critical'' in English are sometimes downgraded to ``Low'' or ``Not Urgent,'' a pattern effectively absent from human annotations.

At the same time, translation also causes the model to systematically favor \emph{Medium}, \emph{High}, or \emph{Critical} classifications, reflecting an overly cautious response strategy. Among scenarios labelled ``Medium'' in English, only $48.5\%$ retain that label upon translation, while $51.5\%$ shift to ``High.'' While such caution may appear desirable in crisis settings, these inconsistencies raise concerns about whether the model consistently captures the underlying urgency of the scenario.

\paragraph{Implications for crisis triage.}
Our findings suggest that automated pipelines based on current state-of-the-art LLMs are not yet suitable for the automatic triage of real-life crisis scenarios. When queried in most languages, LLMs adopt cautious strategies that yield practically ineffective outputs for urgency classification. While a potential solution involves translating all inputs into English before querying the model, this approach remains vulnerable to cascading errors caused by mistranslations or subtle semantic shifts in urgency, as discussed in the preceding sections.

\subsection{Limitations and Future Work}

The findings of this study should be interpreted within the context of several inherent limitations. First, the \dataset{} dataset is composed of synthetic scenarios rather than authentic crisis communications. While this approach allows for a controlled exploration of LLM performance on a simplified classification task, it may not fully capture the complex nuances and linguistic unpredictability characteristic of real-world emergencies. We anticipate that real-world data would present a significantly greater challenge to both translation and classification pipelines than the task addressed in this work.

Additionally, the use of the distilled version of the \texttt{NLLB-200} model likely constrained the translation quality across the evaluated languages. Because this smaller architecture was used, it may have contributed to the performance degradation observed during the urgency assessment phase.

Similarly, this study utilized Llama 3.2 as the primary LLM for classification. Future research should therefore evaluate a broader spectrum of state-of-the-art LLMs to more precisely isolate how different models and varying levels of translation accuracy influence the variance in urgency perception.

Furthermore, a comprehensive analysis of the correlation between translation scores and urgency shifts was not feasible within the current scope. Such an analysis would require a larger volume of parallel data for languages that also possess human annotations to serve as a reliable reference point.

The profiles of our annotators also represent a limitation, as they are not professional crisis responders or subject matter experts. Although our results suggest a robust common understanding of urgency among general users, their perceptions might differ from those of trained crisis actors.

Finally, we recognize that a more rigorous data creation process in the future could involve crisis responders to develop semi-synthetic datasets. Despite these efforts toward better simulation, the ideal foundation for research in this domain remains the collection of authentic crisis data accompanied by high-quality human translations to serve as a definitive gold standard for computational modeling.

\section{Conclusion}
\label{sec:conclusion}

This work examined the suitability of state-of-the-art machine translation systems and LLMs for crisis communication, specifically focusing on the preservation and assessment of urgency. Our results demonstrate that current open-source models, including those explicitly designed for multilingual translation, exhibit substantial limitations in crisis-domain settings, particularly for low- and medium-resource languages. More critically, we show that translation and prompting choices can systematically distort perceived urgency, and that LLM-based urgency classification remains highly unstable across different linguistic contexts.

As a first approach in this domain, we acknowledge that this research serves as a foundational work-in-progress. The study is currently limited by its reliance on a single LLM architecture (Llama 3.2), the use of a distilled translation model (NLLB-200), and a synthetic dataset that, while useful for initial testing, may not capture the full complexity of real-world crisis data. 

These findings raise serious concerns about the deployment of general-purpose language technologies in high-stakes crisis response pipelines without targeted evaluation. For the ISCRAM community, our work underscores the necessity for crisis-aware benchmarks, evaluation criteria, and model design practices that prioritize communicative intent and reliability.

Ensuring that automated systems preserve urgency is essential for building resilient, trustworthy socio-technical systems capable of supporting effective crisis preparedness and response. Finally, we recommend close collaboration between the AI development research community and the crisis informatics community to ensure that future computational progress directly addresses the real-world challenges faced by crisis responders.

\section{Acknowledgments}
This work is supported by the National Science Foundation under the award CIRC 2346334. The authors thank anonymous reviewers, whose suggestions helped improve this article, and Will Lewis, Fei Xia, and Haotian Zhu for their feedback.

\printbibliography

\newpage
\appendix
\section{Annotation Prompt}
\label{app:promptfull}
We explore different prompt designs to describe the task and each of the urgency categories, trying out different punctuation marks to signal the classes. 

\begin{lstlisting}[style=promptstyle, caption={Urgency Classification Prompt}]
Classify the following sentence into one of the 6 levels of urgency in the context of a crisis: Critical, High, Medium, Low, Very Low, Not Urgent.

To help you understand the task, here are some examples for each category in the context of covid 19:

Critical: 'All non-essential movement must cease immediately, a sund all public gatherings are strictly prohibited due to a critical rise in cases overwhelming healthcare capacity.'
High: 'Due to a significant surge in cases, it is now mandatory to wear a medical-grade mask in all public indoor spaces, effective immediately.'
Medium: 'We urge citizens to maintain a physical distance of at least one meter from others in public spaces to mitigate ongoing transmission risks.'
Low: "We encourage all eligible individuals to receive their COVID-19 booster shots to maintain optimal protection against evolving variants."
Very Low: 'Consider wearing a face mask in crowded indoor settings to further reduce your risk of exposure to respiratory viruses.'
Not Urgent: 'Maintaining good hand hygiene, including frequent washing with soap and water, remains a cornerstone of public health, extending beyond the current pandemic.'

Now, classify this sentence in the context of a crisis: 

[SENTENCE]

What is the level of urgency for this sentence?
\end{lstlisting}

\newpage

\section{Cross-lingual Urgency Divergence Examples}
\label{app:divergence}

Table~\ref{tab:llm-divergence} shows four crisis scenarios selected for their highest label entropy across six languages. Each scenario was translated using \texttt{NLLB-200-1B} and then classified by \texttt{Llama}~3.2 (1B). The labels illustrate the breadth of divergence: a single scenario can span the full range from \emph{Not Urgent} to \emph{Critical} depending solely on the prompt language. \emph{N/C} denotes cases where the model refused to classify or produced output outside the label schema.

\begin{table}[H]
  \centering
  \caption{Cross-lingual urgency divergence for four crisis scenarios ranked by label entropy. \texttt{Llama}~3.2 assigns labels ranging from \emph{Not Urgent} to \emph{Critical} for identical text depending solely on the prompt language. \emph{N/C} = non-classified.}
  \includegraphics[trim={38cm 21cm 38cm 0.7cm}, clip, width=\textwidth]{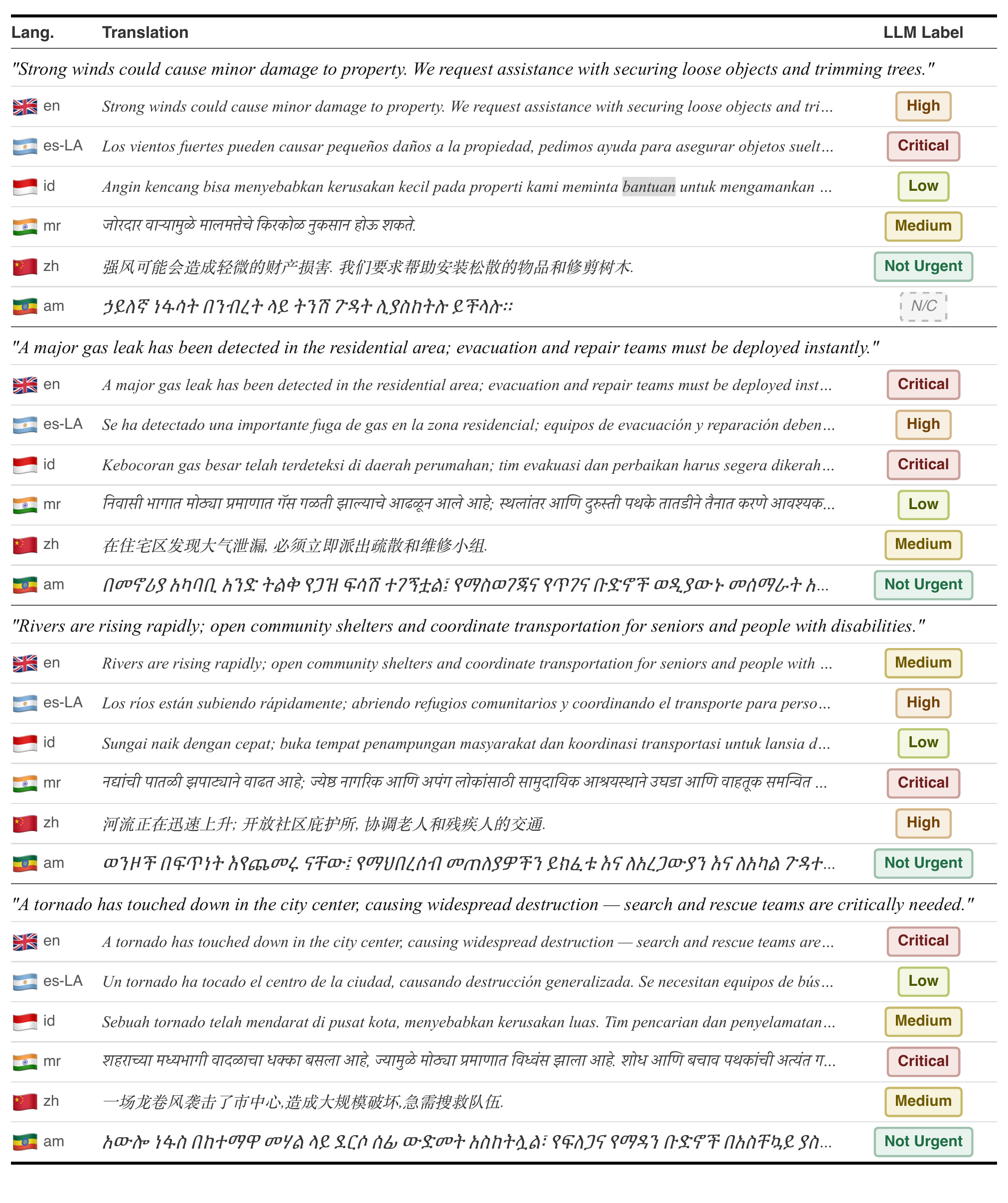}

  \label{tab:llm-divergence}
\end{table}

\newpage

\section{Model Language Support Across TICO-19 Languages}
\label{app:lang-support}

\input{supported-languages}
\input{mt-supported-analysis}
\input{urgency-explained}

\end{document}

%% file: tico-19-mt-table.tex
\newcommand{\best}[1]{\textbf{#1}}
\newcommand{\bestnum}[1]{\multicolumn{1}{r}{\textbf{#1}}}

\begin{table}[!ht]
\centering
\small
\setlength{\tabcolsep}{5.5pt}
\renewcommand{\arraystretch}{1.08}
\caption{Translation Quality ($en \rightarrow X$) Across Geographic Groups. \textbf{Bold} indicates the highest spBLEU score per language. NLLB-1.2B remains the most consistent for generation, while X-ALMA shows surprising strength in specific languages like Somali and Kinyarwanda.}
\label{tab:mt-results-en-x}

\begin{tabular}{@{}lrrrrrrrr@{}}
\toprule
\textbf{Language} & \multicolumn{2}{c}{\textbf{Aya101}} & \multicolumn{2}{c}{\textbf{NLLB-1.2B}} & \multicolumn{2}{c}{\textbf{Llama3.2-1B}} & \multicolumn{2}{c}{\textbf{X-ALMA}} \\
\cmidrule(lr){2-3} \cmidrule(lr){4-5} \cmidrule(lr){6-7} \cmidrule(lr){8-9}
& \textbf{\scriptsize spBLEU} & \textbf{\scriptsize chrF2++} & \textbf{\scriptsize spBLEU} & \textbf{\scriptsize chrF2++} & \textbf{\scriptsize spBLEU} & \textbf{\scriptsize chrF2++} & \textbf{\scriptsize spBLEU} & \textbf{\scriptsize chrF2++} \\
\midrule
\rowcolor{gray!8} \multicolumn{9}{l}{\textit{Africa}} \\
Amharic & 13.23 & 23.80 & \bestnum{25.27} & \bestnum{34.22} & 0.49 & 2.15 & 1.31 & 2.08 \\
Ganda & 7.16 & 17.95 & \bestnum{22.48} & \bestnum{42.66} & 2.14 & 12.71 & 2.48 & 5.85 \\
Hausa & \bestnum{20.36} & \bestnum{41.56} & 18.12 & 36.70 & 1.60 & 12.07 & 2.29 & 3.28 \\
Kinyarwanda & 10.17 & 28.88 & 40.99 & 56.70 & 0.95 & 9.62 & \bestnum{55.92} & \bestnum{70.54} \\
Lingala & 6.20 & 20.41 & \bestnum{21.73} & \bestnum{44.36} & 1.96 & 13.26 & 2.64 & 5.13 \\
Nigerian Fulfulde & 3.70 & 11.33 & \bestnum{44.98} & \bestnum{62.14} & 0.93 & 8.11 & 33.61 & 51.30 \\
Somali & 8.98 & 27.01 & 24.97 & 47.53 & 0.56 & 6.85 & \bestnum{38.24} & \bestnum{51.98} \\
Swahili & \bestnum{29.60} & \bestnum{51.92} & 12.38 & 31.43 & 2.93 & 20.02 & 1.36 & 1.87 \\
Tigrinya & 4.65 & 11.33 & \bestnum{16.76} & \bestnum{26.73} & 0.03 & 1.28 & 2.08 & 2.97 \\
West Cent. Oromo & 4.03 & 18.54 & 34.91 & 53.10 & 0.04 & 0.18 & \bestnum{38.17} & \bestnum{55.24} \\
\addlinespace
\rowcolor{gray!8} \multicolumn{9}{l}{\textit{Southern and Eastern Asia}} \\
Bengali & 22.93 & 39.81 & \bestnum{35.11} & \bestnum{49.29} & 4.46 & 19.94 & 2.67 & 4.02 \\
Burmese & 18.19 & 35.62 & \bestnum{51.79} & \bestnum{70.84} & 0.05 & 0.62 & 49.72 & 68.72 \\
Hindi & \bestnum{29.62} & 47.92 & 29.39 & \bestnum{51.78} & 13.23 & 32.31 & 2.02 & 3.11 \\
Indonesian & 42.87 & 64.22 & \bestnum{48.41} & 48.41 & 26.73 & 50.52 & 27.68 & 44.66 \\
Marathi & 14.46 & 34.60 & \bestnum{23.73} & \bestnum{48.92} & 1.29 & 8.04 & 2.66 & 9.75 \\
Nepali & \bestnum{24.74} & \bestnum{44.30} & 24.19 & 40.16 & 2.98 & 16.86 & 1.57 & 2.26 \\
Standard Malay & \bestnum{41.63} & \bestnum{63.46} & 25.37 & 44.60 & 16.59 & 41.11 & 25.41 & 44.45 \\
Tagalog & 32.34 & 55.05 & \bestnum{47.88} & \bestnum{67.97} & 5.37 & 25.21 & 0.98 & 2.65 \\
Urdu & 22.40 & 41.17 & \bestnum{36.94} & 36.94 & 2.22 & 15.99 & 1.80 & 2.82 \\
Simp. Chinese & 31.06 & 29.25 & \bestnum{33.54} & \bestnum{53.47} & 21.27 & 18.73 & 6.04 & 18.34 \\
\addlinespace
\rowcolor{gray!8} \multicolumn{9}{l}{\textit{West, Eastern, and Central Asia}} \\
Arabic (MSA) & 24.93 & 40.80 & 40.38 & 54.30 & 5.29 & 19.80 & \bestnum{41.46} & \bestnum{54.23} \\
Dari & \bestnum{17.38} & \bestnum{36.99} & 3.23 & 15.52 & 1.33 & 5.95 & 1.27 & 2.64 \\
Kurdish (Kurmanji) & 17.38 & 36.87 & \bestnum{54.69} & \bestnum{72.73} & 0.38 & 3.08 & 53.81 & 71.48 \\
Kurdish Sorani & 1.97 & 2.90 & \bestnum{20.82} & \bestnum{42.94} & 0.30 & 4.17 & 2.27 & 9.82 \\
Russian & 30.05 & 48.18 & \bestnum{32.69} & 32.69 & 13.70 & 29.91 & 2.85 & 15.01 \\
Southern Pashto & \bestnum{19.04} & 35.10 & 16.30 & \bestnum{40.16} & 0.25 & 5.00 & 2.39 & 5.22 \\
Western Persian & 27.39 & 44.50 & \bestnum{58.07} & \bestnum{73.20} & 5.04 & 20.05 & 56.47 & 71.55 \\
\addlinespace
\rowcolor{gray!8} \multicolumn{9}{l}{\textit{Latin America}} \\
Brazilian Portuguese & \bestnum{46.99} & \bestnum{64.65} & 32.04 & 49.26 & 38.12 & 57.99 & 27.19 & 43.54 \\
Latin American Spanish & \bestnum{46.80} & \bestnum{65.15} & 5.63 & 22.84 & 36.75 & 57.10 & 1.89 & 3.14 \\
\addlinespace
\rowcolor{gray!8} \multicolumn{9}{l}{\textit{Europe}} \\
French & 35.86 & 55.37 & \bestnum{37.22} & 37.22 & 27.28 & 48.27 & 36.29 & \bestnum{52.68} \\

\midrule
\textbf{Mean} & 22.38 & 38.38 & \best{32.40} & \best{48.40} & 7.42 & 18.27 & 16.59 & 26.23 \\
\textbf{Std Dev} & 13.20 & 16.33 & 13.91 & 13.68 & 11.23 & 17.58 & 20.35 & 26.69 \\
\textbf{Wins (spBLEU)} & 9 & --- & \textbf{18} & --- & 0 & --- & 4 & --- \\
\bottomrule
\end{tabular}
\end{table}

\begin{table}[!ht]
\centering
\small
\setlength{\tabcolsep}{5.5pt}
\renewcommand{\arraystretch}{1.08}
\caption{Translation Quality ($X \rightarrow \text{en}$) Across Geographic Groups. \textbf{Bold} indicates the highest spBLEU score per language. NLLB-1.2B shows robust performance in African and Southeast Asian languages, while Llama3.2-1B and X-ALMA exhibit significant failure modes in low-resource scripts.}
\label{tab:mt-results}
\begin{tabular}{@{} l rr rr rr rr @{}}
\toprule
\textbf{Language} & \multicolumn{2}{c}{\textbf{Aya101}} & \multicolumn{2}{c}{\textbf{NLLB-1.2B}} & \multicolumn{2}{c}{\textbf{Llama3.2-1B}} & \multicolumn{2}{c}{\textbf{X-ALMA}} \\
\cmidrule(lr){2-3} \cmidrule(lr){4-5} \cmidrule(lr){6-7} \cmidrule(lr){8-9}
& \textbf{\scriptsize spBLEU} & \textbf{\scriptsize chrF2++} & \textbf{\scriptsize spBLEU} & \textbf{\scriptsize chrF2++} & \textbf{\scriptsize spBLEU} & \textbf{\scriptsize chrF2++} & \textbf{\scriptsize spBLEU} & \textbf{\scriptsize chrF2++} \\
\midrule
\rowcolor{gray!8} \multicolumn{9}{l}{\textit{Africa}} \\
Amharic & 28.44 & 50.53 & \bestnum{36.75} & 56.70 & 1.39 & 9.84 & 1.47 & 2.80 \\
Ganda & 27.47 & 44.60 & \bestnum{38.97} & 57.70 & 5.60 & 18.39 & 1.51 & 14.50 \\
Hausa & \bestnum{30.62} & 49.32 & 30.39 & 48.05 & 4.93 & 18.34 & 1.07 & 11.95 \\
Kinyarwanda & 24.69 & 45.48 & \bestnum{40.82} & 61.37 & 3.39 & 16.10 & 22.83 & 38.43 \\
Lingala & 19.75 & 37.90 & \bestnum{37.40} & 54.17 & 2.44 & 15.54 & 1.29 & 14.48 \\
Nigerian Fulfulde & 13.69 & 27.84 & \bestnum{45.84} & 63.08 & 4.94 & 17.00 & 1.41 & 14.42 \\
Somali & 15.96 & 31.69 & \bestnum{32.00} & 51.39 & 2.74 & 13.70 & 1.65 & 16.39 \\
Swahili & \bestnum{38.47} & 58.46 & 19.58 & 34.78 & 15.68 & 35.73 & 1.45 & 13.73 \\
Tigrinya & 26.68 & 47.10 & \bestnum{32.94} & 52.03 & 1.86 & 7.90 & 1.65 & 15.95 \\
West Cent. Oromo & 19.80 & 39.25 & \bestnum{54.82} & 71.48 & 2.95 & 15.37 & 0.90 & 11.97 \\
\addlinespace
\rowcolor{gray!8} \multicolumn{9}{l}{\textit{Southern and Eastern Asia}} \\
Bengali & 39.51 & 60.06 & \bestnum{51.13} & 68.42 & 16.86 & 37.95 & 0.94 & 4.52 \\
Burmese & 29.69 & 51.55 & \bestnum{57.76} & 73.17 & 0.61 & 4.52 & 0.76 & 2.51 \\
Hindi & \bestnum{44.28} & 64.12 & 39.12 & 56.87 & 25.73 & 46.65 & 7.05 & 11.52 \\
Indonesian & 46.08 & 65.60 & \bestnum{56.56} & 73.00 & 34.94 & 56.11 & 40.44 & 60.76 \\
Marathi & \bestnum{32.83} & 55.36 & 31.43 & 49.74 & 12.79 & 32.93 & 2.15 & 6.80 \\
Nepali & \bestnum{42.53} & 62.56 & 36.81 & 57.48 & 10.27 & 28.87 & 8.20 & 17.57 \\
Standard Malay & \bestnum{48.53} & 67.25 & 43.81 & 63.15 & 30.24 & 50.72 & 39.86 & 56.92 \\
Tagalog & 54.51 & 70.52 & \bestnum{62.47} & 75.73 & 28.37 & 46.23 & 0.61 & 2.39 \\
Urdu & \bestnum{33.26} & 55.69 & 32.31 & 58.44 & 16.63 & 38.63 & 2.42 & 4.85 \\
Simp. Chinese & 25.45 & 51.76 & \bestnum{48.84} & 64.79 & 13.06 & 36.67 & 14.35 & 37.06 \\
\addlinespace
\rowcolor{gray!8} \multicolumn{9}{l}{\textit{West and Central Asia}} \\
Arabic (MSA) & 36.03 & 58.60 & \bestnum{46.31} & 66.21 & 21.13 & 44.69 & 10.06 & 22.60 \\
Dari & \bestnum{34.39} & 56.70 & 5.89 & 19.88 & 18.51 & 39.94 & 1.17 & 13.84 \\
Kurdish (Kurmanji) & 34.29 & 54.59 & \bestnum{54.56} & 70.92 & 4.27 & 17.80 & 1.95 & 16.37 \\
Kurdish Sorani & 27.85 & 50.20 & \bestnum{36.29} & 56.59 & 1.23 & 13.09 & 2.06 & 16.49 \\
Russian & 35.97 & 57.73 & 43.44 & 62.57 & 28.25 & 51.69 & \bestnum{53.51} & 69.36 \\
Southern Pashto & \bestnum{33.70} & 55.46 & 32.41 & 51.92 & 3.66 & 20.31 & 6.26 & 12.84 \\
Western Persian & 35.81 & 57.58 & \bestnum{57.37} & 73.49 & 23.95 & 46.15 & 8.10 & 16.78 \\
\addlinespace
\rowcolor{gray!8} \multicolumn{9}{l}{\textit{Latin America}} \\
Brazilian Portuguese & \bestnum{50.10} & 68.99 & 43.44 & 62.68 & 43.85 & 63.64 & 0.64 & 6.09 \\
Latin American Spanish & 48.58 & 68.03 & 14.46 & 29.97 & 41.95 & 62.29 & \bestnum{53.06} & 69.31 \\
\addlinespace
\rowcolor{gray!8} \multicolumn{9}{l}{\textit{Europe}} \\
French & 39.49 & 59.09 & \bestnum{44.56} & 63.60 & 34.52 & 54.56 & 40.99 & 58.30 \\
\midrule
\textbf{Mean} & 33.36 & 53.69 & \textbf{41.40} & \textbf{59.79} & 14.15 & 31.52 & 10.03 & 19.34 \\
\textbf{Std Dev} & 9.94 & 10.97 & 12.59 & 13.08 & 13.11 & 18.06 & 16.14 & 19.08 \\
\textbf{Wins (spBLEU)} & 10 & --- & \textbf{19} & --- & 0 & --- & 2 & --- \\
\bottomrule
\end{tabular}
\end{table}

%% file: supported-languages.tex
Table~\ref{tab:lang-support} lists the 30 TICO-19 evaluation languages alongside their official support status for each model. \checkmark~indicates the language is officially supported; $\sim$ indicates the language is not officially supported but a proximate language group was assigned for \texttt{X-ALMA}; \texttimes~indicates no official support. \texttt{Aya-101} treats both Kurdish varieties as a single macro-language (\texttt{kur}), covering Kurmanji and Sorani without distinction, whereas \texttt{NLLB} distinguishes them as separate codes (\texttt{kmr\_Latn} for Kurmanji and \texttt{ckb\_Arab} for Sorani). For Tagalog, \texttt{Aya-101} provides support via Filipino (\texttt{fil}), which is standardized on Tagalog (\texttt{tgl}). \texttt{Llama}~3.2 officially supports only 8 languages (English, German, French, Italian, Portuguese, Hindi, Spanish, and Thai) and does not officially distinguish language varieties or dialects within those categories. Among the TICO-19 evaluation languages, only Hindi, French, Brazilian Portuguese, and Latin American Spanish fall within its official support.

\begin{table}[!ht]
\centering
\small
\setlength{\tabcolsep}{5.5pt}
\renewcommand{\arraystretch}{1.12}
\caption{TICO-19 evaluation languages and official support status per model: \texttt{X-ALMA}, \texttt{Aya-101}, \texttt{NLLB}, and \texttt{Llama}~3.2. \checkmark~indicates official support; $\sim$~indicates proximate group assignment for \texttt{X-ALMA}; \texttimes~indicates no official support. Group ID refers to the \texttt{X-ALMA} module group.}
\label{tab:lang-support}
\begin{tabular}{@{}lcccccc@{}}
\toprule
\textbf{Language} & \textbf{TICO-19} & \textbf{X-ALMA} & \textbf{Aya-101} & \textbf{NLLB} & \textbf{Llama3.2} & \textbf{Group ID} \\
\midrule
\rowcolor{gray!13} \multicolumn{7}{l}{\textit{Africa}} \\
Amharic            & \texttt{am}     & \texttimes & \checkmark & \checkmark & \texttimes & 8 \\
Ganda              & \texttt{lg}     & \texttimes & \texttimes & \checkmark & \texttimes & 8 \\
Hausa              & \texttt{ha}     & \texttimes & \checkmark & \checkmark & \texttimes & 8 \\
Kinyarwanda        & \texttt{rw}     & \texttimes & \texttimes & \checkmark & \texttimes & 8 \\
Lingala            & \texttt{ln}     & \texttimes & \texttimes & \checkmark & \texttimes & 8 \\
Nigerian Fulfulde  & \texttt{fuv}    & \texttimes & \texttimes & \checkmark & \texttimes & 8 \\
Somali             & \texttt{so}     & \texttimes & \checkmark & \checkmark & \texttimes & 8 \\
Swahili            & \texttt{sw}     & \texttimes & \checkmark & \checkmark & \texttimes & 8 \\
Tigrinya           & \texttt{ti}     & \texttimes & \texttimes & \checkmark & \texttimes & 8 \\
West Cent.\ Oromo  & \texttt{om}     & \texttimes & \texttimes & \checkmark & \texttimes & 8 \\
\midrule[0.3pt]
\rowcolor{gray!13} \multicolumn{7}{l}{\textit{Southern and Eastern Asia}} \\
Bengali            & \texttt{bn}     & \texttimes & \checkmark & \checkmark & \texttimes & 7 \\
Burmese            & \texttt{my}     & \texttimes & \checkmark & \checkmark & \texttimes & 6 \\
Hindi              & \texttt{hi}     & \checkmark & \checkmark & \checkmark & \checkmark & 8 \\
Indonesian         & \texttt{id}     & \checkmark & \checkmark & \checkmark & \texttimes & 4 \\
Marathi            & \texttt{mr}     & \checkmark & \checkmark & \checkmark & \texttimes & 7 \\
Nepali             & \texttt{ne}     & \checkmark & \checkmark & \checkmark & \texttimes & 7 \\
Standard Malay     & \texttt{ms}     & \checkmark & \checkmark & \checkmark & \texttimes & 4 \\
Tagalog            & \texttt{tl}     & \texttimes & \checkmark & \checkmark & \texttimes & 4 \\
Urdu               & \texttt{ur}     & \checkmark & \checkmark & \checkmark & \texttimes & 7 \\
Simplified Chinese & \texttt{zh}     & \checkmark & \checkmark & \checkmark & \texttimes & 6 \\
\midrule[0.3pt]
\rowcolor{gray!13} \multicolumn{7}{l}{\textit{West, Eastern, and Central Asia}} \\
Arabic (MSA)       & \texttt{ar}     & \checkmark & \checkmark & \checkmark & \texttimes & 8 \\
Dari               & \texttt{prs}    & \texttimes & \texttimes & \checkmark & \texttimes & 8 \\
Kurdish (Kurmanji) & \texttt{ku}     & \texttimes & \checkmark & \checkmark & \texttimes & 8 \\
Kurdish (Sorani)   & \texttt{ckb}    & \texttimes & \checkmark & \checkmark & \texttimes & 8 \\
Russian            & \texttt{ru}     & \checkmark & \checkmark & \checkmark & \texttimes & 3 \\
Southern Pashto    & \texttt{ps}     & \texttimes & \checkmark & \checkmark & \texttimes & 8 \\
Western Persian    & \texttt{fa}     & \checkmark & \checkmark & \checkmark & \texttimes & 8 \\
\midrule[0.3pt]
\rowcolor{gray!13} \multicolumn{7}{l}{\textit{Latin America}} \\
Brazilian Portuguese   & \texttt{pt-BR}  & \checkmark & \checkmark & \checkmark & \checkmark & 2 \\
Latin American Spanish & \texttt{es-LA}  & \checkmark & \checkmark & \checkmark & \checkmark & 2 \\
\midrule[0.3pt]
\rowcolor{gray!13} \multicolumn{7}{l}{\textit{Europe}} \\
French             & \texttt{fr}     & \checkmark & \checkmark & \checkmark & \checkmark & 2 \\
\bottomrule
\end{tabular}
\end{table}

%% file: mt-supported-analysis.tex
\section{Additional Machine Translation Analysis}
\label{app:mt-analysis}

This appendix provides additional analyses supporting the machine translation results discussed in Section~\ref{sec:rq1}. In particular, we examine how official language support and geographic region relate to translation quality across the evaluated multilingual systems.

The supplementary figures complement the aggregate results presented in the main paper by showing the full distribution of spBLEU and chrF++ scores, as well as per-language performance variability. Together, these analyses highlight the strong dependence of translation quality on language coverage and reveal substantial disparities across linguistic regions that are not apparent from average benchmark scores alone.

\begin{figure*}[!b]
\centering
\includegraphics[width=.9\textwidth]{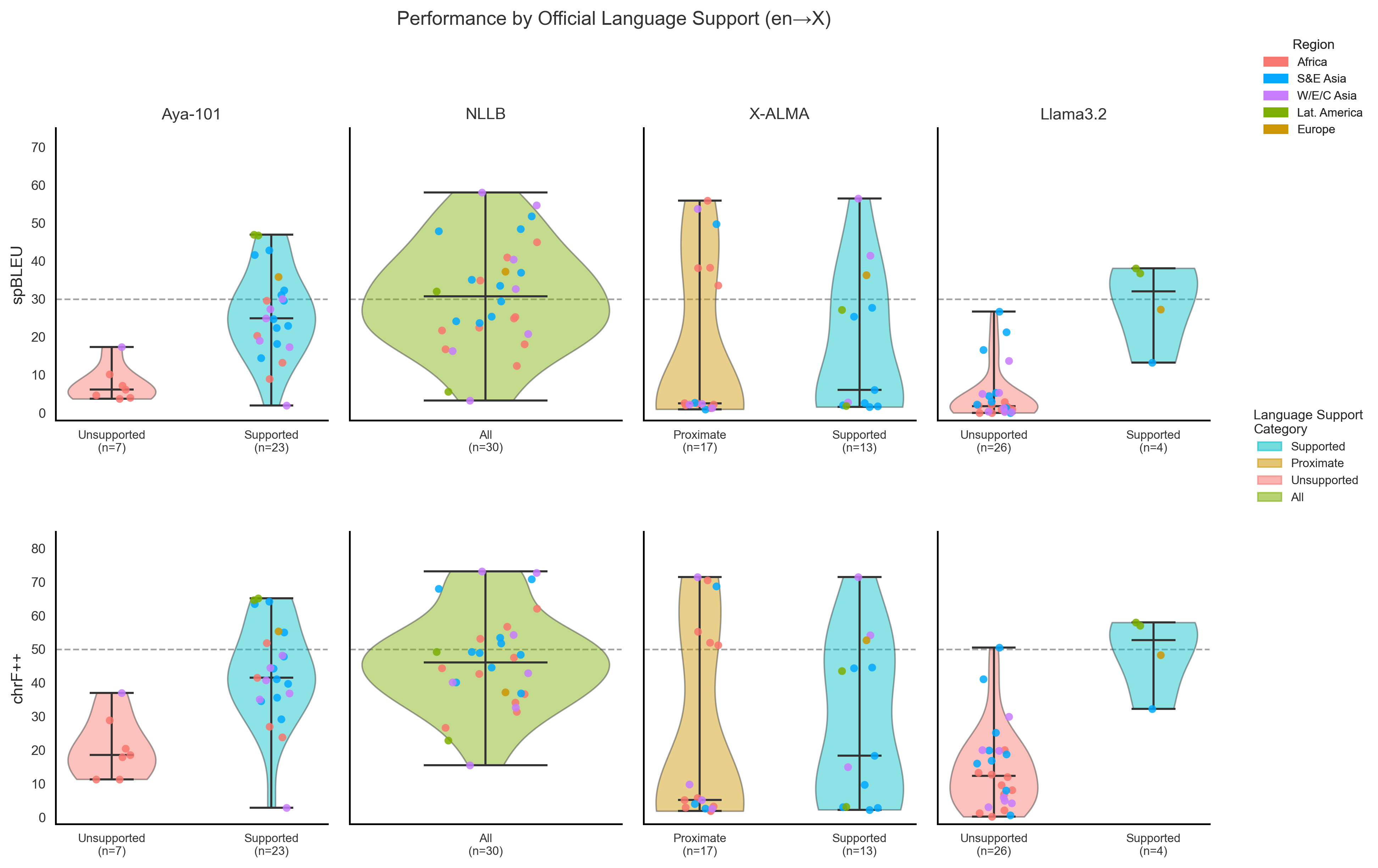}
\caption{
spBLEU and chrF++ performance by official language support category for each MT system in both direction $en \rightarrow X$ direction. Each point represents one language and is colored by geographic region. The dashed line at spBLEU~=~30 marks the approximate threshold for preserving usable semantic content. Supported languages generally achieve substantially higher performance, while unsupported or proximate languages frequently fall below this threshold. Compared to other models, \texttt{NLLB-200} exhibits more consistent performance across regions and fewer catastrophic failures.
}
\label{fig:support_violin}
\end{figure*}

\begin{figure*}[!ht]
\centering
\includegraphics[width=\textwidth]{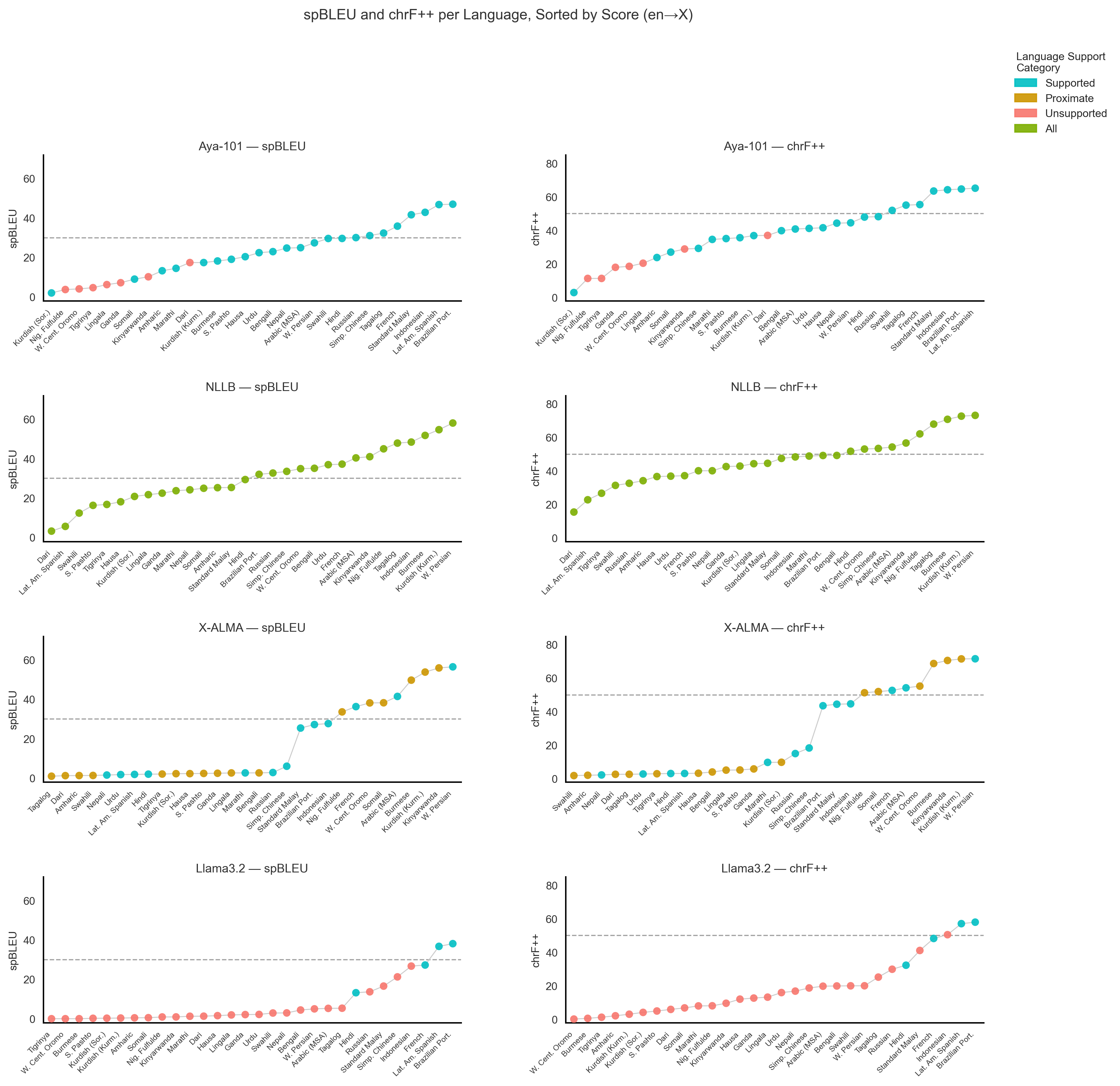}
\caption{
Per-language spBLEU and chrF++ scores for each MT system in the $en \rightarrow X$ direction, sorted by performance. Colors indicate official language support categories. The figure highlights the long-tail behavior hidden by aggregate averages: several multilingual systems exhibit abrupt performance collapse for specific languages despite strong results elsewhere.
}
\label{fig:support_dotplot}
\end{figure*}

%% file: urgency-explained.tex
\newpage

\section{Transition Matrix Construction and Color Encoding}
\label{app:transition-matrices}

In this appendix we describe how the transition matrices shown in Figure~\ref{fig:combinedChangesHALLM} were constructed and how their color scales should be interpreted in more detail.

\paragraph{Shared methodology.}
Both matrices compare urgency labels assigned under two conditions: (i)~when a scenario is presented in English and (ii)~when the same scenario is presented in a non-English language. For each scenario, we compute a cross-tabulation where rows correspond to the English label and columns correspond to the translated-language label. Rows are normalized independently so that each row sums to 100\%, allowing the matrices to be interpreted as conditional transition distributions.

Cell color encodes a \emph{distance-weighted disagreement} score. Cells on the main diagonal correspond to perfect agreement and therefore remain white regardless of frequency. Darker cells indicate disagreements that are both frequent and farther from the diagonal (e.g., shifts from \emph{Critical} to \emph{Not Urgent}). Formally, the color intensity for a cell is defined as:

\begin{equation}
  c_{ij} = \frac{p_{ij} \times |i-j|}{(L_{\max}-L_{\min}) \times 100}
\end{equation}

where $p_{ij}$ is the row-normalized percentage for cell $(i,j)$, and urgency labels are represented as integer levels from 0 (\emph{Not Urgent}) to 5 (\emph{Critical}).

\paragraph{LLM transition matrix.}
The LLM matrix aggregates predictions from \texttt{Llama}~3.2 across the \dataset{} multilingual dataset. Languages in which the model produced valid labels for fewer than $80\%$ of the scenarios were excluded from the analysis, leaving 14 qualifying languages (including English). For each scenario, urgency scores across the qualifying non-English languages were averaged and rounded to the nearest category, producing a single translated-language label per scenario. These averaged labels form the column axis of the matrix, while the English predictions form the row axis. The resulting matrix exhibits strong off-diagonal saturation, indicating that the same scenario may receive substantially different urgency assessments depending solely on the prompt language.

\paragraph{Human annotation transition matrix.}
The human matrix is constructed from annotations collected from 8 graduate-student annotators across four languages: Spanish, Greek, Bengali, and Hindi (two annotators per language). For each language group, annotators evaluated part of the dataset in English and part in their native language, such that all scenarios received annotations under both conditions. English-language scores were averaged to obtain the reference label for each scenario, while translated-language scores were averaged to obtain the corresponding non-English label. Compared to the LLM matrix, the human matrix remains concentrated near the diagonal, indicating that disagreements are generally limited to adjacent urgency categories and that large-magnitude shifts are rare.